\documentclass[%
 reprint,
 amsmath,amssymb,
 aps,
]{revtex4-2}

\usepackage{epsfig}
\usepackage{amsmath}
\usepackage{times}
\usepackage{siunitx}
\usepackage{braket}
\usepackage{booktabs}
\usepackage{chemist}
\usepackage{mathtools}
\usepackage{float}
\usepackage{comment}
\usepackage{appendix}
\usepackage{graphicx}
\usepackage{dcolumn}
\usepackage{bm}

\begin{document}

\preprint{APS/123-QED}

\title{Towards Multicellular Biological Deep Neural Nets\\ Based on Transcriptional Regulation}

\author{Sihao Huang \\\href{mailto:sihao@mit.edu}{sihao@mit.edu}}
\affiliation{Department of Physics, Massachusetts Institute of Technology}

\begin{abstract}
Artificial neurons built on synthetic gene networks have potential applications ranging from complex cellular decision-making to bioreactor regulation. Furthermore, due to the high information throughput of natural systems, it provides an interesting candidate for biologically-based supercomputing and analog simulations of traditionally intractable problems. In this paper, we propose an architecture for constructing multicellular neural networks and programmable nonlinear systems. We design an artificial neuron based on gene regulatory networks and optimize its dynamics for modularity. Using gene expression models, we simulate its ability to perform arbitrary linear classifications from multiple inputs. Finally, we construct a two-layer neural network to demonstrate scalability and nonlinear decision boundaries and discuss future directions for utilizing uncontrolled neurons in computational tasks.
\end{abstract}

\maketitle

\section{\label{sec:level1}Introduction}
Simple neural networks built using synthetic gene networks, metabolic circuits \cite{pandi_koch_voyvodic_soudier_bonnet_kushwaha_faulon_2019}, and DNA strand displacement \cite{qian_winfree_bruck_2011} have demonstrated learning models in single cells. These circuits can potentially be used in a bio-processing environment to adapt to contextual perturbations and in building robust systems that can respond to patient physiology in complex ways. Scaling these elements up to multi-layer neural networks also opens up the possibility of performing arbitrary, higher-dimensional classifications, such as those demonstrated using genetic logic circuits. This approach also has the added advantage of being dynamically re-configurable, allowing for adaptive control, optimization, and backpropagation using the same design. \\

Despite the relatively slow response time of synthetic gene networks, the density and scalability of this system also makes it a potential candidate for information processing. Namely, assuming the same tissue density and metabolic rate as a human, up to 500 trillion cells can fit within a cubic meter. This fully supported and self-assembling unit would represent an equivalent energy consumption of 1700 W \cite{mcmurray_soares_caspersen_mccurdy_2014}. Short of full cellular control and readout, the complex nonlinear dynamics and tunable coupling between cells in a 2D or 3D array can potentially be used for random graph traversals. By mixing various neuron types and varying the tissue density, the model can also be of interest in condensed matter physics, random matrix theory, and the investigation of critical phase phenomenon. \\

However, there are a number of major roadblocks ahead of scaling up cell-based neural networks, including model training and interfacing, intercellular signaling, network connectivity, and physical construction. The first challenge which we would like to address is the design of a modular, fully capable neuron. Here, we build a deterministic model of a single artificial neuron based on transcriptional regulation to demonstrate arbitrary linear classifications. We also compose the system into a multi-layered neural network capable of performing nonlinear classifications. 

\section{\label{sec:level1}System-Level Design}

\begin{figure}[b]
\includegraphics[width=0.5\textwidth]{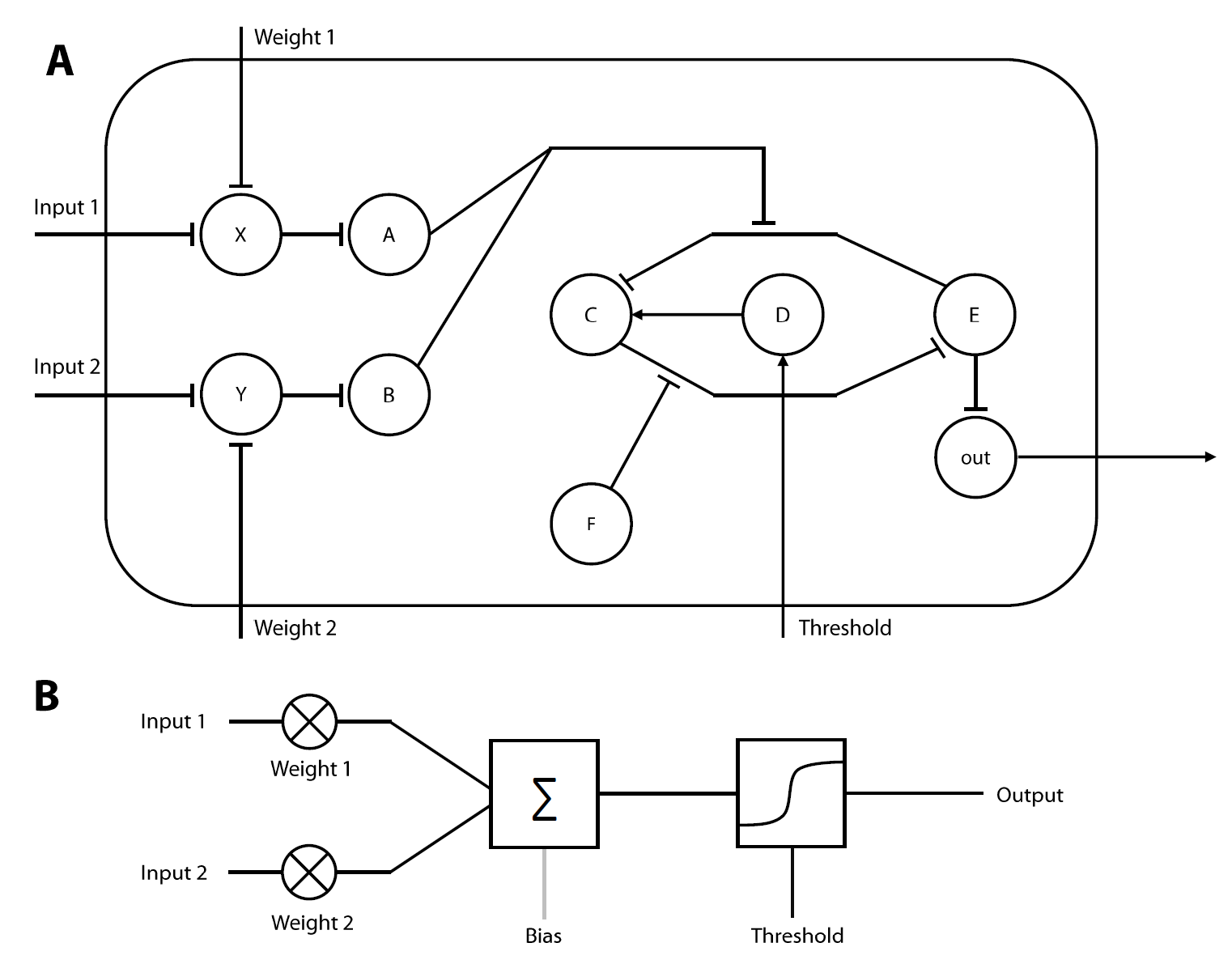}
\caption{\label{fig:epsart} \textbf{A}: High-level gene regulatory network (GRN) for implementing a general two-input neuron. \textbf{B}: Schematic representation of the artificial neuron.}
\label{sys_schematic}
\end{figure}

A genetic circuit is first designed to emulate the two-input artificial neuron in Figure \ref{sys_schematic}. The system takes in two inputs, multiplies them with weights $w_1$ and $w_2$ and takes their sum. This value is passed into a sigmoidal transfer function, given by $S(x) = \frac{1}{1 + e^{- (x - T)}}$, with a variable threshold T. The output is high if the sum of the inputs, multiplied by the weights, surpass the threshold value. Together, this forms a general linear classifier for two inputs,

\begin{equation}
    w_1 x + w_2 y = T.
\end{equation}

By varying $w_1$, $w_2$, and $T$, we are able to draw arbitrary decision boundaries on a 2D plane. \\

The high-level gene regulatory system is described in \ref{sys_schematic}.\textbf{A}. There are seven functionally important nodes, creating a two-input artificial neuron with adjustable weights and a variable threshold. The nodes labeled X and Y consist of variable concentration inputs, which are passed to nodes A and B after being multiplied by an arbitrarily variable weight. The multiplication is accomplished by an intermediate signaling node that is activated by weights 1 and 2, effectively acting as an adjustable attenuator. Nodes A and B both induce C, summing the value of the two inputs. \\

Nodes C and E mutually repress and act as a toggle switch. This provides the sigmoidal transfer function of the artificial neuron. Node E is followed by the output node, which is controlled under a separate operator sequence to enable signal matching. With no neuronal input, node E is high by default as the repression of E by C is sequestered by a constitutive bias supplied from F. However, if the weighted sum of nodes A and B is sufficiently high, the repression of node C by node E is overcome, and the bistable circuit transitions E into a low state. One can then engineer the threshold value of the neuron by adding an inducer to node C, which is controlled by an externally supplied small molecule.  

\begin{figure*}
\includegraphics[width=1\textwidth]{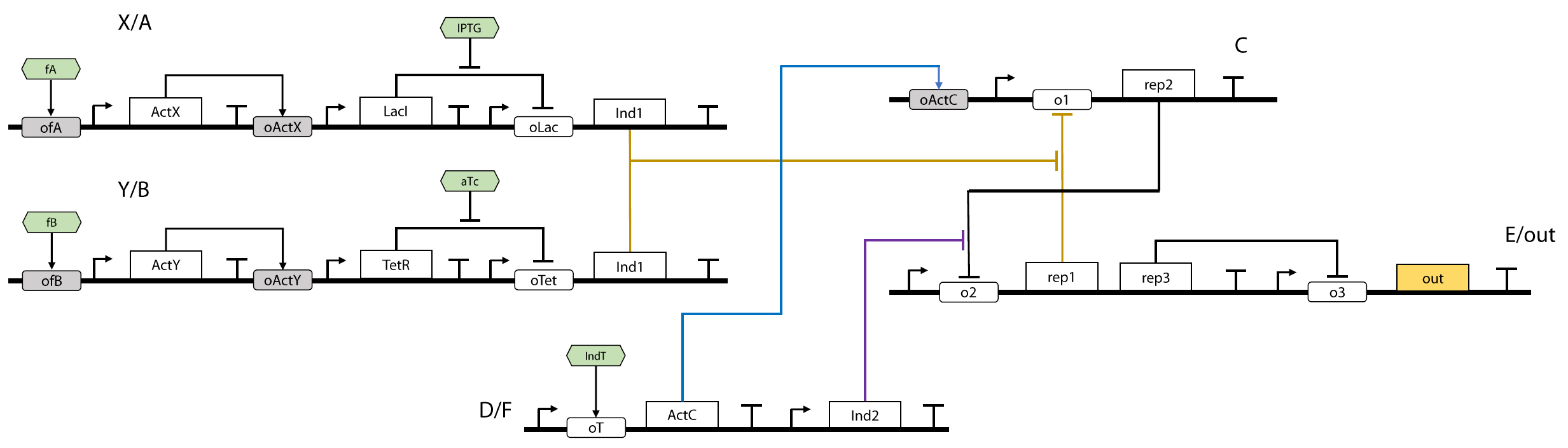}
\caption{\label{fig:wide}Gene regulatory network for implementing a general two-input artificial neuron. Square elements indicate protein coding sequences and rounded elements indicate operators. Arrows designate activation while lines with blunt ends designate repression.}
\end{figure*}

\section{\label{sec:level1}Gene Network Implementation}

Our goal is to produce a generic design that does not require specialized promoters or repressors and enables straightforward implementation. We arrived at the gene regulatory network shown in Figure 2 after an iterative design process. Each part, labeled A to E, maps roughly to the nodes in \textbf{(2)}. The network has a number of explicit state variables which are listed in Table 1.\\

The dynamics of this network is modeled via of a set of coupled nonlinear differential equations with Hill-like repression and activation terms. We note that stochastic processes and biophysical delays are not accounted for in these equations. An analysis of noise processes on a similar circuit can be found in reference \cite{bates_blyuss_alsaedi_zaikin_2015}. 

\begin{table}
\caption{\label{tab:table1}List of modeled state variables}
\begin{tabular}{ l r }
                                    \toprule
     State variable & Description\\ \midrule
     IPTG & Input 1 ($x$)\\
     aTc & Input 2 ($y$)\\ 
     IndT & Inducer for threshold control \\
     fA & Inducer for ActX ($w_1$) \\
     fB & Inducer for ActY ($w_2$) \\
     ActX & Activator for LacI expression \\
     ActY & Activator for TetR expression \\ \midrule
     LacI & Lac repressor protein\\
     TetR & Tetracycline repressor protein\\ 
     Ind1 & Inducer for input summation \\
     Ind2 & Inducer for basal activation of node C \\
     rep1 & Node C repressor protein \\
     rep2 & Toggle switch cross-repression protein\\
     rep3 & Output protein repressor \\
     out & Output or signaling intermediate\\ \bottomrule
  \end{tabular}
\end{table}

\subsection{\label{sec:level2}Modeling Methodology}

As the system is not sensitive to race conditions, the precise timing of state transitions is not critical. We opt to assume that mRNA dynamics are fast. Specifically, this means we assume that mRNAs immediately reach the steady-state concentrations set by their production and degradation terms, as degradation is quick compared to protein dynamics. \\

A similar assumption is used for the binding and dissociation of transcription factors to and from the operator sites, wherein the equilibrium concentration is reached quickly. As a result, each binding reaction can be described by a single rate constant. The repression and activation of each gene can then be modeled using Hill terms from these rates. \\

Sequestration reactions are modeled by second-order mass-action kinetics with a forward and reverse rate constant. We assuming that binding is irreversible, i.e. the degradation of the repressor proteins occur before being the dissociation of the bound inducer. Finally, we model the signaling proteins with first-order degradation processes determined by their respective concentrations.

\subsection{\label{sec:level2}Mechanistic Equations}

The input to the circuit is supplied via $IPGT$ and $aTc$, which bind to $LacI$ and $TetR$ respectively. The $LacI$ and $TetR$ proteins are controlled by activators $ActX$ and $ActY$, which are in turn controlled by inducers $fA$ and $fB$. \\

The activation of the $LacI$ and $TetR$ genes by the presence of $ActX$ and $ActY$ is modeled by Hill terms with cooperativity  $n_{ActX}$ and $n_{ActY}$, which scale the constitutive production rates $k_{prod, LacI}$ and $k_{prod, TetR}$. We note that the Hill approximation holds only when the promoter or repressor are in large excess compared to their binding sites, which is generally true in system with a low plasmid copy number. These rates are distinct as each gene is structurally different and may have different copy numbers if parameter tuning is required. The natural, first order degradation of $ActE$ is also accounted for in the rate equation.

\begin{align}
    \frac{d[ActX]}{dt} = k_{prod, ActX} \frac{[fA]^{n_{fA}}}{K_{fA}^{n_{fA}} + [fA]^{n_{fA}}} - k_{deg, ActX}[ActX]
\end{align}

\begin{eqnarray}
    \frac{d[LacI]}{dt} = k_{prod, LacI} \frac{[ActX]^{n_{ActX}}}{K_{ActX}^{n_{ActX}}
    + [ActX]^{n_{ActX}}} \nonumber\\
    - k_{deg, LacI}[LacI] - k_{seq, LacI}[LacI][IPTG]
\end{eqnarray}

\begin{align}
    \frac{d[ActY]}{dt} = k_{prod, ActY} \frac{[fB]^{n_{fB}}}{K_{fB}^{n_{fB}} + [fB]^{n_{fB}}} - k_{deg, ActY}[ActY]
\end{align}

\begin{eqnarray}
    \frac{d[TetR]}{dt} = k_{prod, TetR} \frac{[ActY]^{n_{ActY}}}{K_{ActY}^{n_{ActY}} +  [ActY]^{n_{ActY}}} \nonumber\\ - k_{deg, TetR}[TetR] - k_{seq, TetR}[TetR][aTc]
\end{eqnarray}

This mechanism allows multiplicative analog weighting of the inputs. The steady state concentration of $ActX \cdot degA$ and $ActY \cdot degB$ can be described by functions $f_x([fA])$ and $f_y([fB])$, which is then multiplied by $[IPTG]$ and $[aTc]$ respectively through concentrations of $LacI$ and $TetR$. \\

If $IPTG$ or $aTc$ are low, $Ind1$ will be produced from strands A and B, which then sequesters $rep1$, pushing the toggle switch towards a high state.

\begin{eqnarray}
    \frac{d[Ind1]}{dt} = k_{prodA} \frac{K_{LacI}^{n_{LacI}}} {K_{LacI}^{n_{LacI}} + [LacI]^{n_{LacI}}} \nonumber\\
    + k_{prodB} \frac{K_{TetR}^{n_{TetR}}} {K_{TetR}^{n_{TetR}} + [TetR]^{n_{TetR}}} - k_{degInd1}[Ind1]
\end{eqnarray}

This is counterbalanced by a basal expression of $Ind2$. When $[Ind1]$ is high enough, the action of $Ind2$ can be overcome, thus driving the output of the neuron high. However, when it drops below a certain threshold, $Ind2$ again becomes dominant, acting as a reset mechanism for the artificial neuron. 

\begin{align}
    \frac{d[Ind2]}{dt} = k_{prodInd2} - k_{degInd2}[Ind2]
\end{align}

Ind1 sequesters $rep1$, which enables the expression of $rep2$ on strand C as the first part of the toggle circuit. The gene is controlled by a hybrid promoter consisting of $oActC$ and $o1$. This can be modeled by a Hill-like activation term with cooperativity $n_{ActC}$ and a Hill-like repression term with cooperativity $n_{rep1}$. For the circuit to function as intended, we want $rep2$ to be primarily controlled by $o1$, but also activated by $ActC$ to shift the bistable threshold. Such a control logic can be accomplished by either multiplying the two Hill terms or adding them, however, in practice, the former will only function for low Hill coefficients, i.e. when the switching regime is mostly linear. Therefore, we opt to implement an $OR(ActC, NOT(rep1))$ condition by summing the two terms. Depending on the specific biological implementation, the hybrid promoter can also be broken up into two strands if needed. 

\begin{eqnarray}
    \frac{d[rep2]}{dt} = k_{prodC} \frac{K_{rep1}^{n_{rep1}}} {K_{rep1}^{n_{rep1}} + [rep1]^{n_{rep1}}} \nonumber\\
    + k_{prodC} \frac{[ActC]^{n_{ActC}}} {K_{ActC}^{n_{ActC}} + [ActC]^{n_{ActC}}} \nonumber\\
    - k_{seqrep2}[rep2][Ind2] - k_{degrep2}[rep2]
\end{eqnarray}

$rep1$ forms the second part of the cross-repression circuit. Its expression is controlled by the same operator $o2$ as $rep3$, which then controls the output protein.

\begin{eqnarray}
    \frac{d[rep1]}{dt} = \frac{d[rep3]}{dt} = k_{prodE} \frac{K_{rep2}^{n_{rep2}}} {K_{rep2}^{n_{rep2}} + [rep2]^{n_{rep2}}} \nonumber\\
    - k_{seqrep1} [rep1] [Ind1] - k_{degrep1} [rep1]
\end{eqnarray}

By separating the output, we are able to tune its expression levels by either selecting a different repressor or by adding promoters to the sequence.  

\begin{align}
    \frac{d[out]}{dt} = k_{prodout} \frac{K_{rep3}^{n_{rep3}}} {K_{rep3}^{n_{rep3}} + [rep3]^{n_{rep3}}} 
    - k_{degout}[out]
\end{align}

Finally, $ActC$, which biases node C and drives it towards a high state, is expressed through the introduction of $IndT$. This sets the threshold value of the neuron.

\begin{align}
    \frac{d[ActC]}{dt} = k_{prodD} \frac{[IndT]^{n_{IndT}}} {K_{IndT}^{n_{IndT}} + [IndT]^{n_{IndT}}}
    - k_{degActC}[ActC]
\end{align}

Here, we note that the specific mechanism through which the expression of $ActC$ is controlled is not particularly important, as long as it is stable and easily variable via external means. While a repressor and a small molecule inducer can also be used, we model it using a basic promoter as it is not critical in determining the dynamics of the artificial neuron.

\section{\label{sec:level1}Analysis and Optimization}

Optimizing the system involve selecting the appropriate repressors, promoters, and copy numbers to establish the right rate constants and cooperativity for robust behavior. Signaling proteins and their associated transcription factor must be chosen such that steady-state concentrations are bound between 100 to 2000 molecules per cell. The lower bound is set by thermodynamic limits as stochastic processes start to significantly degrade the signal-to-noise ratio of the system below this level. The upper limit is set by energetic resources and shared transcription/translation machinery, the precise value of which can be determined by detailed modeling or through inference from natural expression rates in the cell. As the circuit uses standard transcription factors and does not require exceptionally high degrees of cooperatively, it should be feasible to find a complete set of biological components to realize the design experimentally. \\

The concentrations given in this analysis are expressed in nanomolars, where [1nM] is approximately equal to one molecule per cell, assuming that the volume of a cell is $10^{-15}L$ (this is roughly the size of chassis organisms such as E. coli). Literature values are used for $LacI$, $TetR$, $IPTG$, and $aTc$. \\

\subsection{\label{sec:level2}Stability Analysis}

\begin{figure}[b]
\includegraphics[width=0.5\textwidth]{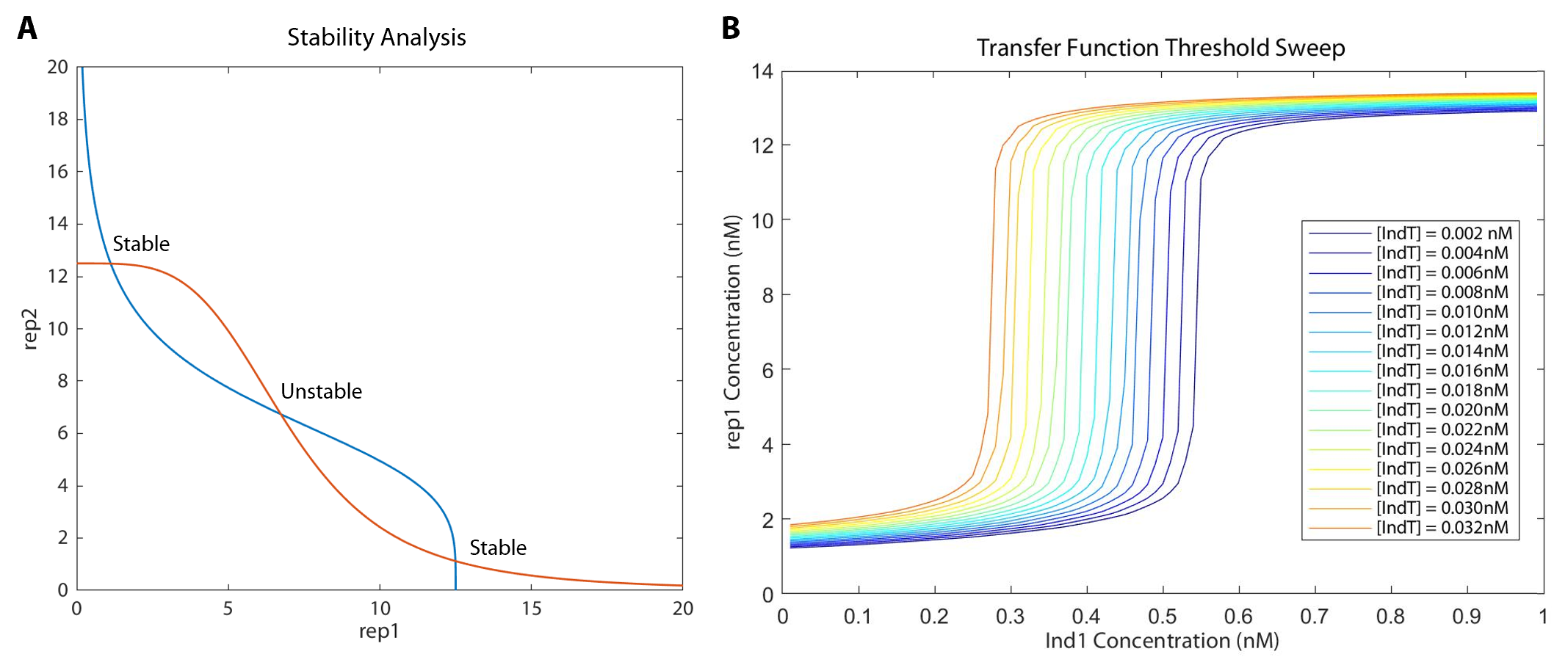}
\caption{\label{fig:epsart} \textbf{A}: Nullcline analysis of the toggle network. \textbf{B}: Change in the neuron transfer function as $IndT$ concentration is swept. Note that the system is bistable for lower and higher concentrations of $Ind1$ and transitions sharply in the intermediate state.}
\end{figure}

We chose to build the model starting from the most constrained module to maximize the design space for the rest of the gene network. Since the bistable toggle switch is the most sensitive to tuning parameters, it is critical select the right parameters for this part of the circuit. \\

The nullclines for the two mutually repressing nodes is similar in structure to the toggle switch proposed in \cite{gardner_cantor_collins_2000}. It can be described by a set of two Hill functions,

\begin{eqnarray}
    k_{prodC} \frac{K_{rep1}^{n_{rep1}}} {K_{rep1}^{n_{rep1}} + [rep1]^{n_{rep1}}}- k_{degrep2}[rep2] = 0
\end{eqnarray}

\begin{eqnarray}
    k_{prodE} \frac{K_{rep2}^{n_{rep2}}} {K_{rep2}^{n_{rep2}} + [rep2]^{n_{rep2}}} - k_{degrep1} [rep1] = 0.
\end{eqnarray}

The design parameters for the toggle switch must be chosen such that the two stable points land within positive $[rep1]$ and $[rep2]$ values while maintaining maximum separation between the curves. This ensures that the bistability is less prone to stochastic switching, and increases the dynamic range between the on and off states of the system. Furthermore, slope of the transfer function, which is determined by the Hill coefficients of $rep1$ and $rep2$ binding, sets the bounds for signal matching to downstream neurons. As a result, a higher binding cooperativity in the two repressors will enable greater network depths and more room for error when building deeper neural networks. The optimized stable points are shown in panel A above. \\


\subsection{\label{sec:level2}Threshold Control}

\begin{figure}
\includegraphics[width=0.5\textwidth]{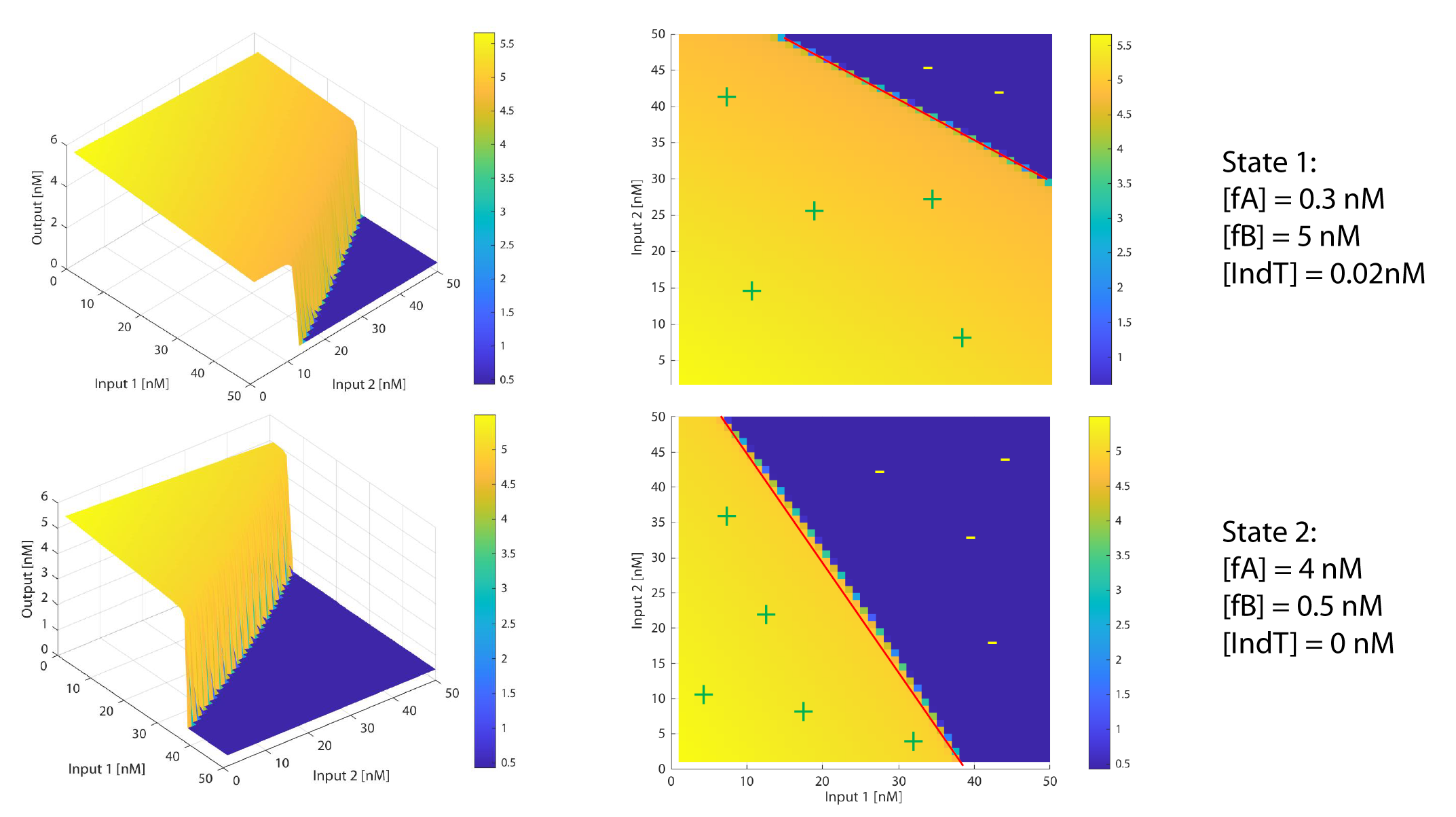}
\caption{\label{fig:epsart} Examples of programmable linear decision boundaries using a single GRN-based neuron. States 1 and 2 are achieved by tuning the inputs weights $[fA]$ and $[fB]$ as well as the threshold $[IndT]$.}
\end{figure}

The threshold of the artificial neuron is controlled via node D, which directly expresses node C. By changing this activation strength via an external inducer $IndT$, we are able to change the $Ind1$ concentration at which the neuron transitions into the high state. In panel $\textbf{B}$, this is seen as a linear translation of the transfer function. \\

It is important to note that we want to maximize the tuning range of this threshold because it essentially sets the size of the classification plane, defined by the inputs, in which the neuron can operate. By increasing the tuning range, we effectively increase the dynamic range of this circuit by raising the concentrations higher above the noise floor.

\subsection{\label{sec:level2}Linear Classification}

Putting the circuit together, we can create a 2-dimensional linear classifier by tuning the weight and threshold parameters of the GRN. The plots in Figure 4 are generated by sweeping input 1 ($IPTG$) and input 2 ($aTc$) concentrations, and recording the concentration of $out$ after steady state has been reached. \\

By varying $[fA]$ and $[fB]$, we are able to change the slope of the curve, and by changing $[IndT]$, we are able to translate the decision boundary. This enables us to reach an area ranging from $0 nM$ to $50 nM$ on each input. We note that this graph shows a number of desirable characteristics. First, the transition between the high and low states is sharp. This minimizes the region in the classifier in which the output state is ambiguous, making signal matching more robust when neurons are chained downstream. Second, the input and output concentrations are well separated, with almost no leakage of $[out]$ in the low state.\\

As long as we can find orthogonal repressors, we can increase the number inputs in the neuron by tiling the design of strands A and B. This way, the classifier can be can generalize to $n$-dimensional space, where the decision boundary is a $n-1$-dimensional hyperplane that divides the state space into two.

\section{\label{sec:level1}Multi-stage Network}

The next step is to scale the circuit to perform nonlinear classifications using a two-layer neural network. Biologically, this is significant because it would establish the modularity of the neurons. Functionally, the implementation of nonlinear classifications also open the system up to more complex physics, as well as the emulation of logical circuits, amongst other machine learning tasks. \\

Physically implementing this system would require multiple repressors for the six weights and three threshold values. However, given that the diffusion of intermediate signaling proteins outside of the cell is sufficiently small, the rest of the circuitry can be reused without orthogonality constraints. \\

In the circuit depicted in Figure 5, $IPTG$ and $aTc$ are shared inputs to neuron 1 and neuron 2. This effectively ties the $x$ and $y$ inputs together, which allow the first two input neurons to act as independent linear classifiers. This architecture enables nonlinear decision boundaries to be drawn on the same 2D-plane. \\

We note that this model has 174 free parameters and uses a set of 30 coupled differential equations. As a result, it is very computationally intensive and represent a huge search space to optimize over. Selective simulations and careful modular design are required to make designing and constructing the circuit tenable.

\begin{figure}
\includegraphics[width=0.5\textwidth]{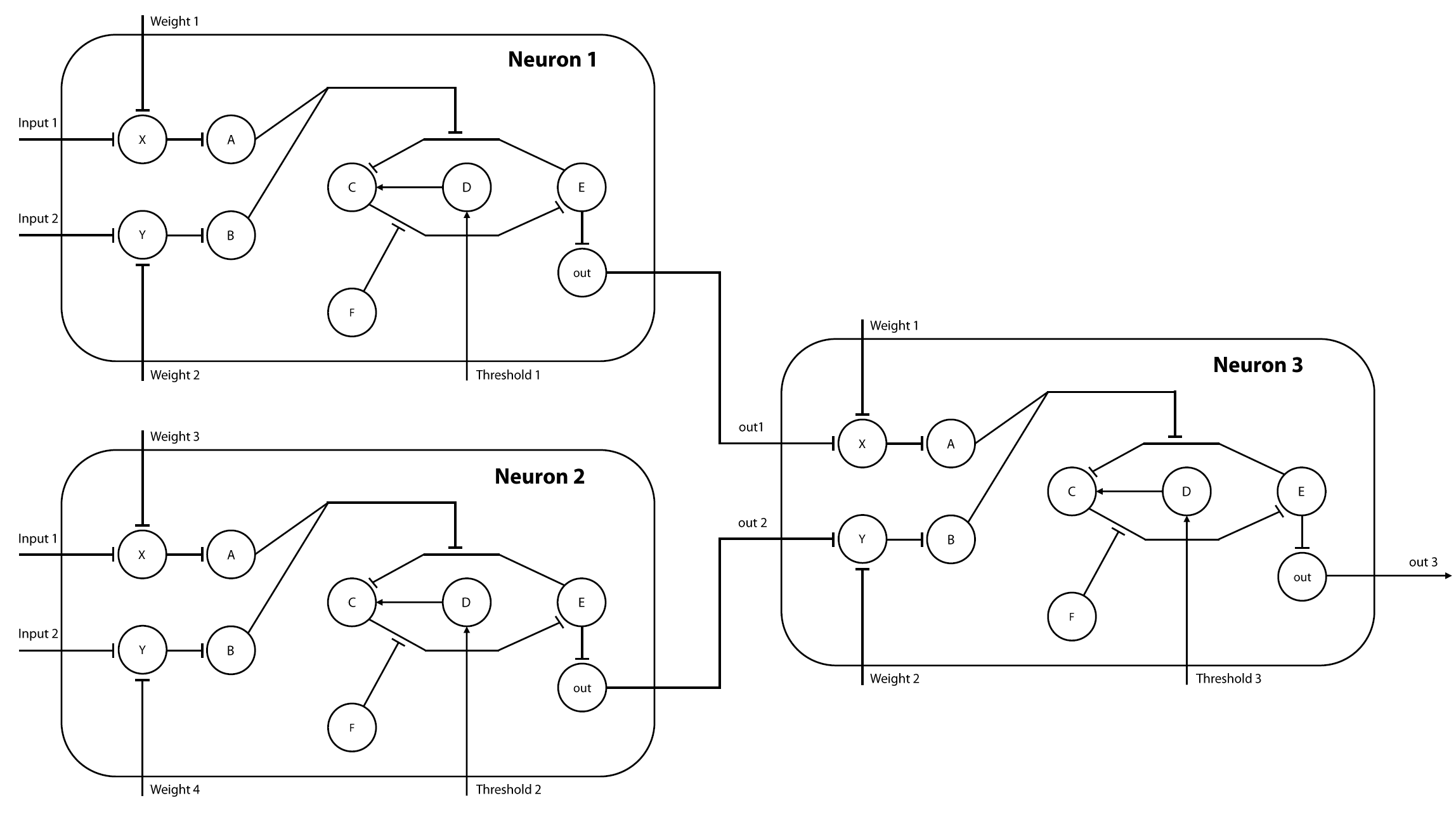}
\caption{\label{fig:epsart} High-level GRN for a two-layer neural network implemented using modular artificial neurons. Neuron 1 and neuron 2 form the input layer, while neuron 3 produces the output.}
\end{figure}

\subsection{\label{sec:level2}Signal Matching}

The multi-neuron design requires precise signal matching between the output of the first layer and the threshold on the second layer. Both $out1$ and $out2$ are expressed at a nearly constant value as a result of the bistability of the system when the neurons are firing. This enables modular design of the neural network, In fact, we can engineer the signal levels to implement logic functions that merge the linear mappings from each layer-one neuron into a nonlinear map on the final output. \\

If the activation threshold for the layer-two neuron is set below the concentrations $[out1]$ or $[out2]$ high individually, but at a sufficient level such that it is relatively noise-resilient, the threshold implements an $OR$ function such that the output neuron fires if either inputs are high. This merges the two mappings, represented by the purple and blue triangles in Figure 6, into a nonlinear classification boundary. Alternatively, we can also implement an $AND$ function if we set the threshold value to be above the concentration of $out1$ and $out2$ alone, but less than the combined value if both neurons fire at the same time. 

\begin{figure}[t]
\includegraphics[width=0.5\textwidth]{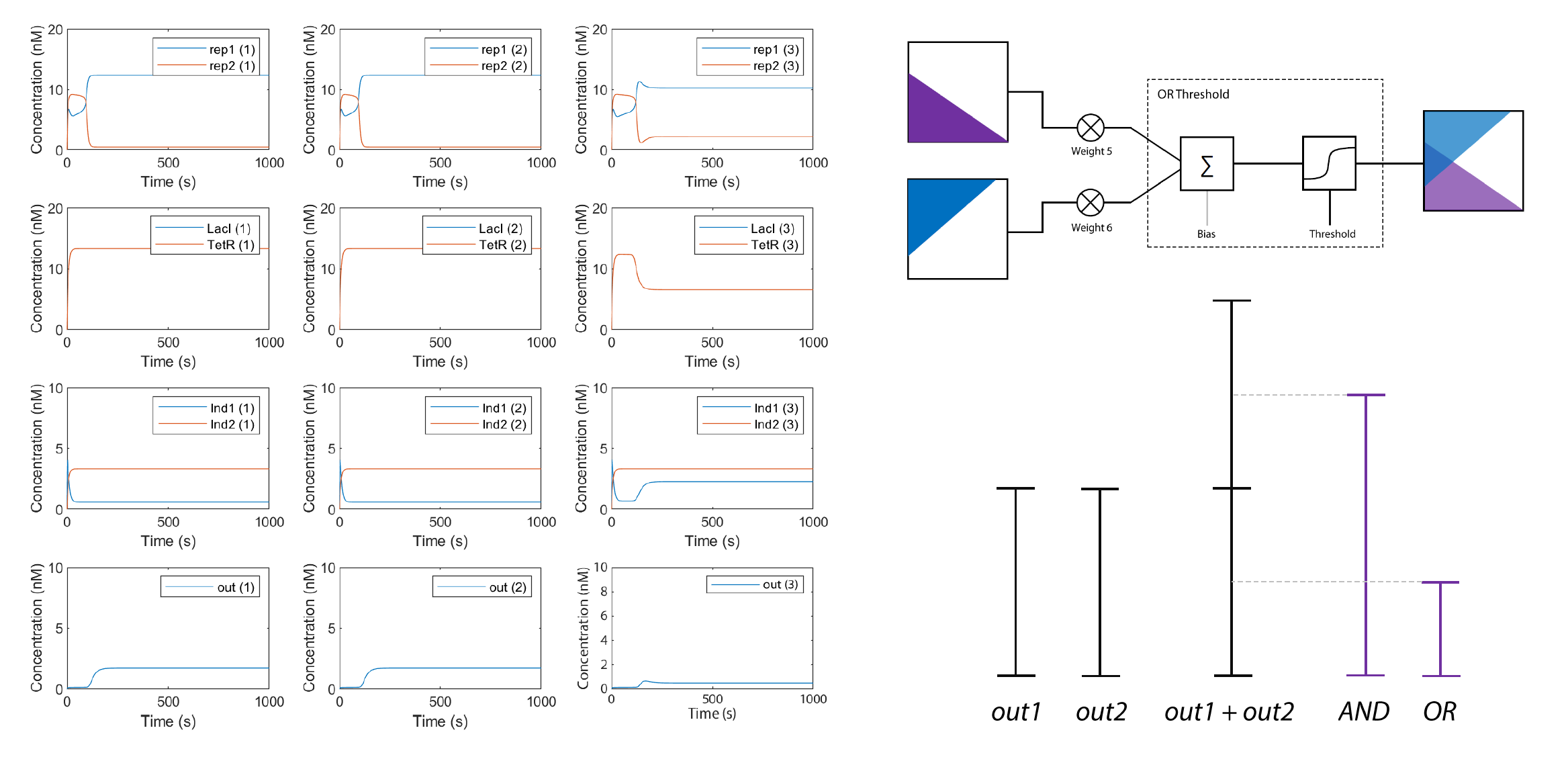}
\caption{\label{fig:epsart} Left panel: transient simulation of a three-neuron network. The first, second, and third column correspond to neurons 1, 2, and 3 respectively. Right, top: a nonlinear decision boundary generated from two linear classifiers. Right, bottom: signal matching thresholds for $OR$ and $AND$ merging.}
\end{figure}

\subsection{\label{sec:level2}Hazards}

In the time-domain simulation in Figure 6, $IPTG$ is introduced into the system at $t = 0$. Due to the basal expression of $Ind2$, the concentration of $rep2$ first rises and then falls as $Ind1$ is expressed from strand A in both neurons 1 and 2. This pushes the neurons towards a high output state. As a result, the concentration of $rep1$ and $rep2$ in neuron 3 also experiences multiple transients during this period. Fortunately, the signal chain acted as a sufficient low-pass filter such that no error state was resulted. \\

In the case of computational applications, hazards can be mitigated by simply reading output after the statement has reached steady-state. However, for therapeutic and biological applications, these transients might result in the erroneous actuation of irreversible components such as killer proteins. In this case, filters can be engineered between each neuron stage to suppress transient behavior. 

\subsection{\label{sec:level2}Programmable Gates}

Using the techniques described above, we can turn the two-layer neural network into any arbitrary two-input gate. This allows us one to construct one gene regulatory network and program it \textit{in-vivo} into different Boolean circuits given sufficient network depth by simply varying the presence of various species in solution. \\

This technique resembles the function of Field Programmable Gate Arrays (FPGAs). While programmable logic has been implemented in platforms such as DNA excision \cite{weinberg_pham_caraballo_lozanoski_engel_bhatia_wong_2017}, neural nets with well-defined transfer functions can also be used for arbitrary threshold sensing. Such a system could enable the rapid development of therapeutic classifiers and so-called smart viruses with weights and thresholds set by the basal expression on user-programmable coding sequences. 

\section{\label{sec:level1}Physical Implementation}

A major hurdle before these artificial neurons could be used for information processing is the physical construction of the network and its interface. \\ 

\begin{figure}[b]
\includegraphics[width=0.5\textwidth]{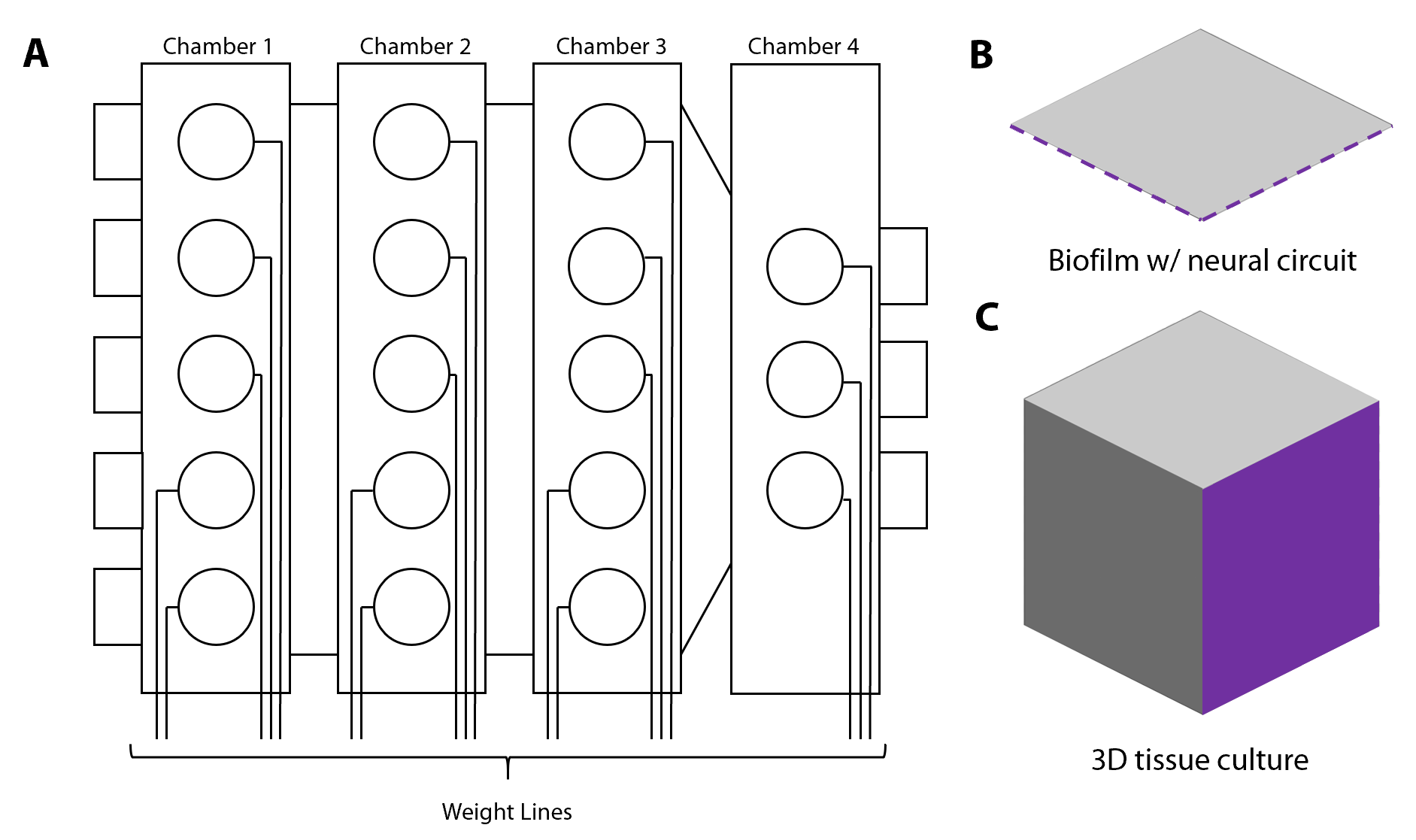}
\caption{\label{fig:epsart} \textbf{A}: Schematic for a microfluidic device that implements a multicellular artificial neural network. \textbf{B}: Biofilms consisting of artificial neurons could be used for information processing. The purple edges indicate control surfaces. \textbf{C}: Similarly, 3D tissue cultures can enable more complex dynamics.}
\end{figure}

A near-term realizable implementation is proposed in Figure 7. The cells are housed in a microfluidic flow cell consisting of multiple staged chambers. Every chamber contains multiple colonies, each under the control of weight lines that changes $[fA]$ and $[fB]$ locally. A constant flow supplies network inputs and moves intermediate species to the next layer. Finally, training is performed \textbf{in-silico} by performing gradient descent on the outputs. \\

A major bottleneck of this system that it would require a large number of orthogonal signaling molecules. Alternatively, we can establish isolation using local concentration gradients. This is the operating principle behind \textbf{B} and \textbf{C}. Specifically, the artificial neuron cells is grown in a biofilm or 3D tissue culture and edge excitations are sent in by changing chemical concentrations at the boundaries. By controlling the coupling between each cell, we can induce random walks in the material to perform inference or graph sampling tasks. \\

Hybrid approaches are also possible with implementations \textbf{B} and \textbf{C}. One can create locally inhomogeneous tissue with synthetic clusters consisting of a small number of orthogonal signaling molecules. These diffusive tissues can implement regional feedback or even contain supervisor-student pairs \cite{nesbeth_2016} to perform backpropagation and supervised learning. Furthermore, diverse cultures with behaviors and coupling terms sampled from a random distribution can be tuned display critical phase phenomenon, opening up the possibility of using techniques from condensed matter physics to study these biological lattices.

\section{\label{sec:level1}Discussion}

We have demonstrated a modular implementation of a general-purpose transcriptional regulation-based neuron that is capable of performing arbitrary linear classifications on a set of input variables. Furthermore, we have shown that these neurons can be composed into a multi-layered neural network which can then act as nonlinear function approximators or as digital circuits that are programmable \textit{in-vivo}. \\

Although there are a number of implementational details that could not be addressed with this study, our simulations have demonstrated the basic feasibility of building deep neural networks using genetic neurons. We made use of basic activator and repressor components which are readily available to synthetic biologists and showed that the circuit could be contained to a reasonable size. \\

We hope that these results would spur further interest in the development of modular, multicellular genetically-based neural networks. While the first step is to find an experimentally realizable version of this neuron, having such a building block would open up many interesting questions in fields ranging from biophysics to machine learning. Techniques such as random mixtures, structured tissue and biofilm cultures, and inhomogeneous subpopulations enable us to engineer complex physics into these systems and potentially harness it to solve computationally interesting problems.

\bibliography{main}

\end{document}